\documentclass{article}

\usepackage{microtype}
\usepackage{graphicx}
\usepackage{subfigure}
\usepackage{booktabs} 

\usepackage[utf8]{inputenc} 
\usepackage[T1]{fontenc}    
\usepackage{url}            
\usepackage{amsfonts}       
\usepackage{nicefrac}       
\usepackage{tabu,multirow}
\usepackage{subfiles}
\usepackage{multicol}
\usepackage{pbox}
\usepackage{makecell}
\usepackage{colortbl}
\usepackage{amsmath}
\usepackage{float}
\usepackage{subfigure}
\usepackage{algorithmic}
\usepackage{hyperref}

\usepackage[accepted]{icml2021}

\icmltitlerunning{Re-understanding Finite-State Representations of Recurrent Policy Networks}

\begin{document}

\twocolumn[
\icmltitle{Re-understanding Finite-State Representations of Recurrent Policy Networks}




\begin{icmlauthorlist}
\icmlauthor{Mohamad H. Danesh}{osu}
\icmlauthor{Anurag Koul}{osu}
\icmlauthor{Alan Fern}{osu}
\icmlauthor{Saeed Khorram}{osu}
\end{icmlauthorlist}

\icmlaffiliation{osu}{Department of EECS, Oregon State University, Corvallis, OR, USA}

\icmlcorrespondingauthor{Mohamad H. Danesh}{daneshm@oregonstate.edu}

\icmlkeywords{Reinforcement Learning, Interpretability}

\vskip 0.3in
]



\printAffiliationsAndNotice{}  

\begin{abstract}
We introduce an approach for understanding control policies represented as recurrent neural networks. Recent work has approached this problem by transforming such recurrent policy networks into finite-state machines (FSM) and then analyzing the equivalent minimized FSM. While this led to interesting insights, the minimization process can obscure a deeper understanding of a machine's operation by merging states that are semantically distinct. To address this issue, we introduce an analysis approach that starts with an unminimized FSM and applies more-interpretable reductions that preserve the key decision points of the policy. We also contribute an attention tool to attain a deeper understanding of the role of observations in the decisions. Our case studies on 7 Atari games and 3 control benchmarks demonstrate that the approach can reveal insights that have not been previously noticed.
\end{abstract}


\section{Introduction}
\label{sec:intro}

What roles do observations and memory play in the decision making of deep policy networks for complex control tasks? 
While such networks have yielded state-of-the-art performance in reinforcement and imitation learning (e.g. \cite{DRLmnih2013playing, DRLhausknecht2015deep, DRLwang2015dueling, DRLvan2016deep,ILho2016generative}), there are limited tools and approaches for giving insight into this question. This is particularly the case for policies represented as recurrent networks, e.g. LSTMs and GRUs \cite{DSLhochreiter1997long, DSLchung2014empirical, DSLcho2014learning}, which condition on high-dimensional memory vectors with no preconceived semantics. Prior works have attempted to gain insight via attention maps over the network input, e.g. \cite{pmlr-v80-greydanus18a, DBLP:journals/corr/abs-1809-06061, gupta2020explain, atrey2020exploratory}, however, this ultimately involves subjective human interpretation of the underlying ``strategic role" of the attended-to elements. In this paper, we develop an analysis approach that reveals such human interpretations can sometimes be highly misleading.  


Our work builds on a recent approach for understanding recurrent policy networks by quantizing memory and observations within an RNN \cite{MISCkoul2018learning}. A quantized RNN is a type of FSM, known as a Moore Machine (MM), which can be visualized and analyzed. In particular, the originally large MMs were algorithmically minimized, resulting in small machines for a variety of domains, which yielded high-level insights. For example, the minimal MM for Atari Pong showed that memory was not actually needed (i.e. a state-action mapping), while for Atari Bowling the MM was an open-loop strategy that ignored observations. 

While analyzing minimal MMs allows for determining global properties, such as memory/observation use or not, we have found it difficult to gain more in-depth insight from minimal MMs. To see this, Figure \ref{minimalMM} (\emph{a}) and (\emph{b}) show minimal MMs for Pong and Bowling and Figure~\ref{frames} shows a set of frames/observations for each MM that all map to a single state in the corresponding MM. For a human, the frames are semantically distinct, whereas the minimization process was able to merge them all into a single state. This is due to the minimization focusing on maintaining logically equivalence to the unminimized MM policy, rather than maintaining any semantic meaning of states. Thus, we have found it very difficult to understand the strategic role of different states in such MMs to gain deeper insight into their decision logic. This experience has led us to the view that minimal MMs are \emph{unlikely} to be a good starting point for gaining a deeper understanding of recurrent policies.




\begin{figure*}[t]
\centering
\includegraphics[width=\textwidth]{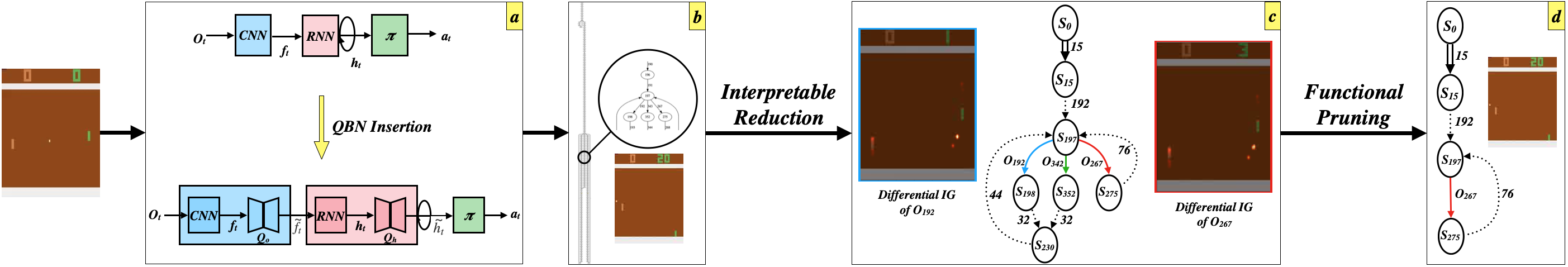}
\caption{Overall approach for Pong. a) \emph{QBN Insertion} (Section~\ref{sec:moore}) discretizes observations and memory of the original RNN (top). b) Resulting \emph{MM} with finite states and observations. c) \emph{Interpretable reductions} (Section~\ref{sec:reduction}) are applied to the MM, yielding a single decision point $s_{197}$ conditioning on observations. \emph{Differential attention} (Section~\ref{sec:attention}) is used to understand decisions in terms of observations. d) \emph{Functional pruning} (Section~\ref{sec:functional}) removes unnecessary branches, leaving an open-loop policy.}
\label{overall_story}
\end{figure*}

Our main contribution is to develop a new approach for RL researchers and advanced practitioners to analyze MM policy representations. We demonstrate that this approach can yield significantly more insight into the decision making of recurrent policies. Rather than start with a minimal MM after quantization, we start with the unminimized MM, which is often quite large, and apply ``interpretable reductions" in order to reduce the visual complexity. The reductions are intentionally simple in order to preserve the inherent decision structure in the machine while effective at compressing the visual representation. For example, one reduction operation replaces a fixed series of non-branching memory states (i.e. a macro) with a single abstract transition. The result is a simplified machine with a fixed set of ``branching states" where the flow of the machine depends on observations. 

In order to help understand the decisions made at branching states, we further develop a new differential attention tool, aiming to identify parts of observations that are most important for selecting one branch over another. Using the tool, we found that often, especially in Atari, the attention was unintuitive from a strategic point of view, which led us to question whether observations were used for strategically important reasons, or whether they were ``arbitrary" branches. This led us to consider the ``functional pruning" reduction to assess that issue. In particular, this operation eliminates all but one branch at a decision point to test whether the observation-conditioned decision among multiple branches was strategically important, or just an artifact of learning a non-compact policy. 

We explore this approach by studying of 7 deterministic Atari games and 3 continuous control environments. For the control tasks, the approach identifies interpretable machines, whose decision points are understandable and strategically meaningful. In contrast, for Atari, the analysis reveals new and surprising insights. Prior works have attempted to understand the most salient pixels for policy decisions in Atari games \cite{pmlr-v80-greydanus18a, DBLP:journals/corr/abs-1809-06061, gupta2020explain}, however, little insight was gained into how that information was used. Our approach reveals that for the Atari policies, observations were not used for ``strategically useful" purposes. In particular, at each state that branches on observations, it was possible to remove all branches, except for one, resulting in an open-loop policy that ignores observations, while maintaining performance. We call such policies, \emph{pruned open-loop policies}, and observe that all of our Atari policies are of this type. Finally, we identify limits of the current approach as problems become more stochastic, which suggests important avenues of future work. 

\section{Related work}

\textbf{Attention Maps.} Attention maps have been used as a tool to identify the most relevant parts of the input with respect to the agent's decision \cite{pmlr-v80-greydanus18a, DBLP:journals/corr/abs-1809-06061, gupta2020explain, atrey2020exploratory}. Perturbation-based attention methods have been investigated to gain insight into learned Atari policies \cite{pmlr-v80-greydanus18a,DBLP:journals/corr/abs-1809-06061, gupta2020explain}, but have been criticized for relying on the application of networks to potentially non-sensical perturbed states \cite{atrey2020exploratory}. This has been partly addressed by using more advanced counterfactual state generation \cite{olson2019counterfactual}. In general, attention approaches produce a ``local explanation" for the decisions made at specific states, which is in contrast to ``global explanations" we primarily focus on in this work. Attention methods are also limited in the type of insights they can provide. For example, while they are applied to recurrent policies, they provide no insight into how memory is used and the strategic role of salient inputs. As we will show, our results suggest that a human's intuition about the role of salient inputs can be highly misleading.   

\begin{figure*}[t]
\begin{center}
\subfigure[]{
\includegraphics[width=0.41\textwidth]{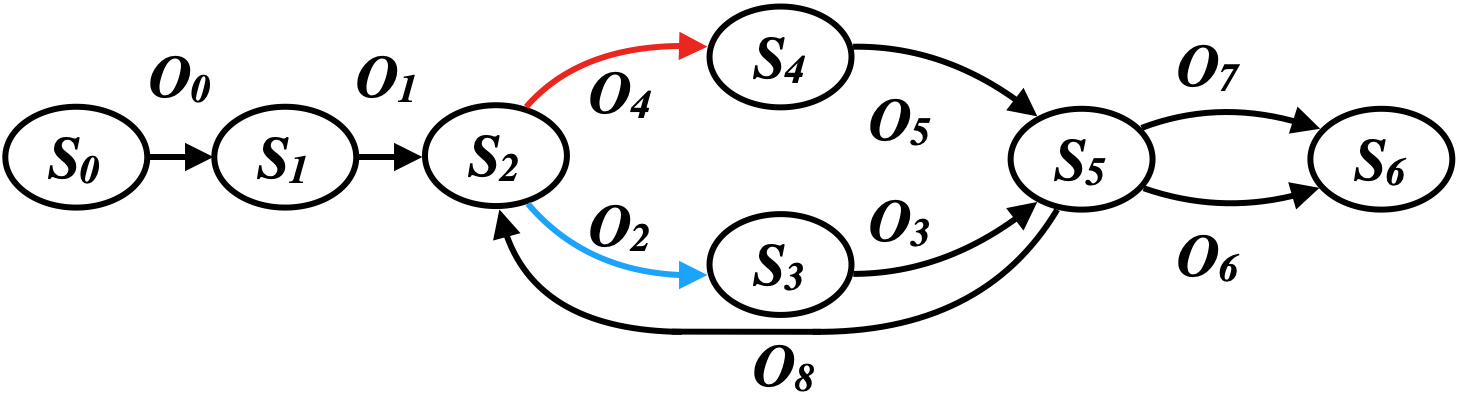}}
\subfigure[] {
\includegraphics[width=0.35\textwidth]{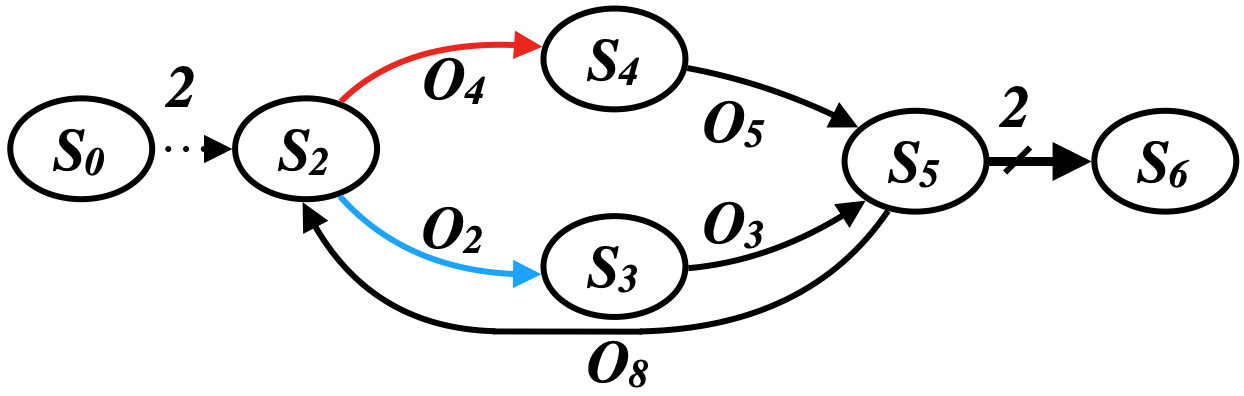}}
\end{center}
\caption{a) An example of an MM with 7 memory states and 9 observations. The initial memory state is $S_0$, and based on the input observation, it goes to the next memory state until it reaches to the memory state $S_5$, then it loops back to $S_2$. In this example, MM has a decision point at $S_2$. b) Interpretable reductions applied to the MM.}
\label{MM-samples}
\end{figure*}

\textbf{Hybrid Architectures.} There have been efforts to induce explanations components in the architecture to make agents implicitly explainable \cite{DBLP:journals/corr/abs-1812-11276, DBLP:journals/corr/abs-1906-02500}. For example, soft attention modules have been used in a recurrent architecture to gain insight into the ``attention" of an agent \cite{DBLP:journals/corr/abs-1906-02500}. While attention has been used in several context for explanation, the soundness of this approach is not clear. In particular, the attention weights are computed from the raw observations and memory, which is ignored in the explanation process. Thus, it is difficult to determine whether the attention patterns carry key strategic information based on other parts of the raw observation, or whether, indeed, it is only the information highlighted by the attention that is key to decisions. Investigating ways to distinguish these two cases is interesting future work. 

\textbf{Extracting Finite-State Machine.} There has been significant prior work on extracting FSMs from RNNs in the context of formal language learning of fixed finite alphabets \cite{FSMtivno1998finite, FSMcechin2003automata, FSMweiss2017extracting, FSMweiss2019learning}. Only recently have there been attempts to learn FSMs for complex RL problems \cite{MISCkoul2018learning} where machine minimization was used to extract high-level insights. Our work builds on \cite{MISCkoul2018learning} and aims to significantly advance the depth of understanding that such methods can provide. 

\section{Recurrent Networks to Moore Machines} \label{sec:moore}

In this section we review \emph{recurrent policy networks (RPNs)} and how they can be converted to \emph{MMs} as illustrated in Figure~\ref{overall_story}a. 
This work is agnostic about how a policy is learned, which, for example, could be via RL, imitation learning, or other training approaches.

{\bf Recurrent Policy Networks (RPNs).} 
An RPN is an RNN policy that, at each time step, is given an observation $o_t$ and outputs an action $a_t$. As illustrated in Figure~\ref{overall_story}a top, during execution, an RPN maintains a continuous-valued hidden memory state $h_t$, which is updated on each transition and influences the action choice $a_t$. Specifically, given current observation $o_t$ and current state $h_t$, an RPN outputs an action $a_t = \pi(h_t)$ where $\pi$ may be a feed-forward network. Then, it updates the state according to $h_{t+1}=\delta(f_t,h_t)$, where $f_t$ is a set of features extracted from $o_t$, for example, using a CNN when observations are images. $\delta$ is the transition function, which is often implemented via different types of gating networks such as LSTMs or GRUs.

{\bf Moore Machines.} Understanding action choices of an RPN is complicated by the memory's high-dimensionality and lack of predefined semantics. Recent work \cite{MISCkoul2018learning} has attempted to address this issue by transforming RPNs to MMs, which allows for visualization and analysis of a finite system. An MM is a finite-state machine defined by a finite set of labeled hidden states $H$, a distinguished initial state $h_0\in H$, a finite set of observation symbols $O$, and a transition function $\Delta : H \times O \rightarrow H$, which returns the next state $h_{t+1}= \Delta(h_t,o_t)$, given the current state and observation symbol. The label associated with each state corresponds to an action. An MM policy initializes the state to $h_0$ and then updates the state as observations arrive and outputs the action associated with each state. 

{\bf Quantized Bottleneck Insertion (QBN Insertion).} We now overview the approach of \emph{Quantized Bottleneck Insertion} \cite{MISCkoul2018learning} for transforming an RPN to an MM policy, which is illustrated in Figure~\ref{overall_story}a bottom. Full details of this approach can be found in the original paper and are not critical to the contributions of this paper. The key components of the approach are \emph{Quantized Bottleneck Networks (QBNs)}, which are simply auto-encoders, for which the encoder produces a quantized latent representation. In this work, bottleneck representation is composed of discrete units with output values in $\{-1, 0, 1\}$. Given a trained RPN, a representative set of RPN trajectories is produced and the resulting sets of hidden states $\{h_t\}$ and observation features $\{f_t\}$ are collected. Next, a hidden-state QBN $Q_h$ and observation QBN $Q_o$ are trained to minimize reconstruction error on the data sets. The encoders of $Q_h$ and $Q_o$ can be viewed as discretizing the state and observation spaces. The trained QBNs are then inserted into the RPN in place of the ``wires" that propagate the continuous memory vector $h_t$ and observation features $f_t$ (see Figure~\ref{overall_story}a). This creates a discrete representation of the hidden states $\hat{h}_t$ and observation features $\hat{f}_t$ in the RPN. In practice, QBNs have reconstruction errors, which may impact the RPN performance. Supervised fine-tuning of the discretized RPN can be used to improve performance via imitation learning with respect to the original RPN. 

Trajectories of the discretized RPN are then run to collect a representative transition set $\{(\hat{h}_t, a_t, \hat{f}_t, \hat{h}_{t+1})\}$, indicating that action $a_t$ was taken in discrete state $\hat{h}_t$ and a transition to $\hat{h}_{t+1}$ was observed when the discrete observation was $\hat{f}_t$. The MM is constructed by creating a transition graph edge for each data tuple. The key parameters relevant to this paper are sizes of the bottlenecks of $Q_h$ and $Q_o$, denoted $N_h$ and $N_o$ respectively. Larger values give more potential to produce finer grained quantization and in turn more states. 

\begin{figure*}[t]
\includegraphics[width=\textwidth]{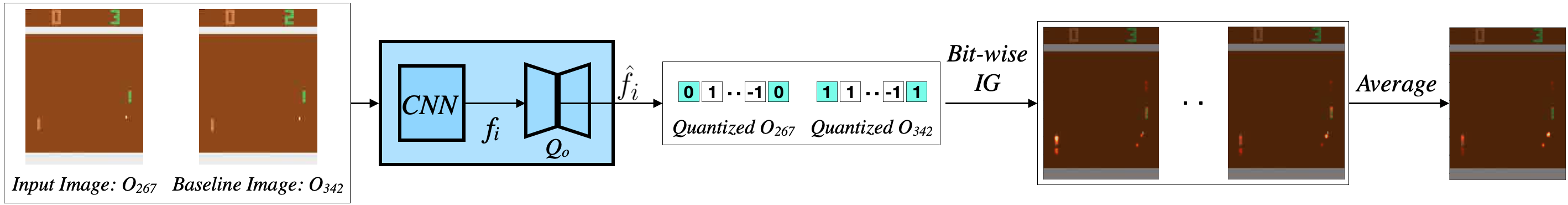}
\caption{Differential Attention Pipeline. The pair of images under comparison result in discrete representations that differ on the highlighted features. We produce an attention map for each of those features using the Integrated Gradient (IG) approach and average the magnitude of the maps for an overall differential attention map.}
\label{fig:attention}
\end{figure*}

\section{Analyzing non-minimal Moore Machines}
RPNs learned for complex problems can result in MMs with large numbers of discrete states and observations. In order to aid understanding, previous work \cite{MISCkoul2018learning} used a standard minimization algorithm \cite{MISCpaull1959minimizing} to produce equivalent minimal state MMs. As described in Section~\ref{sec:intro}, the minimal machines allowed for interesting global insights into the general use of states and observations. However, as also described in Section \ref{sec:intro}, we have found that minimal MMs obscure the decision making behavior, since they compress the original MM structure with no regard for interpretability. For example, the observations mapped to single states often appear to be semantically very different (Figure \ref{frames}), making it difficult to understand the role of states and how observations influence their choices. Thus, in this paper, we start with the unminimized MMs with the aim to gain a more detailed understanding. Below, we describe the steps of our analysis approach. 

\subsection{Interpretable Reductions} \label{sec:reduction}
Figure~\ref{MM-samples}a shows an example MM, which illustrates the main structures we observed in the learned MMs we analyzed. These structures provide several opportunities for simple \emph{interpretable reductions}, which simplify the visualization of an MM without obscuring decision structure. \textit{The reductions are very simple by design and one of our contributions is to notice that this simple set can be effective for interpreting very large MMs.} Note that some of the following reductions may hide some steps, for example long macros with no decision points, to make the visualization manageable. But such simplifications do not essentially influence the completeness of the decision process in MM. The first stage of our approach applies the following interpretable reductions. 

\textbf{Sequence Reduction.} It is common to see long sequences of states with a single observation between consecutive states (e.g. $S_0$ to $S_2$ in Figure~\ref{MM-samples}a). These sequences are open open-loop macros that simply execute a fixed sequence of actions whenever encountered. Examples are $S_0$ and $S_1$ in the sequence of state-transitions from state $S_0$ to $S_2$ that have no branching states. We reduce these sequences to a single ``macro arc" represented as a dotted line with a number indicating its length. 

\textbf{Loop Unrolling Reduction.} There are many loops attached to sequences that are only traversed once when the sequence is visited. These loops increase the visual complexity of the MMs and make it appear as if there is a decision that controls loop exit, when there is not. Thus, we simply unroll such loops before applying sequence reduction. To identify such loops, once the MM is generated, we run the policy once more and count the number of times each node is visited; thus we exactly know how many times a loop is covered.

\textbf{Parallel Reduction.} There are often multiple transition arcs between two states with different observation labels (e.g. between $S_5$ and $S_6$ in Figure~\ref{MM-samples}a). This can also occur for self-transitions. We merge these arcs into an abstract arc labeled by the number of observations. 

\textbf{Startup and Termination Reduction.} In some MMs, there is a period of state transitions that corresponds to a warm-up/termination period in the environment where actions have no impact. If desired, these parts of an MM can be replaced with a macro arc represented by two parallel lines with the number of transitions in that period. This reduction requires minimal human annotation of the steps where episodes ``meaningfully" begin and end, which is usually straightforward. For example, in Atari games, there is usually a warm-up period before actions impact the game and the machines can have arbitrary structure that is not important to the game playing behavior. It is straightforward for a human to mark the time when this warm-up period ends. 

The result of these reductions is shown in Figure~\ref{MM-samples}b and often result in orders of magnitude smaller visualizations, e.g., going from Figure~\ref{overall_story}b to Figure~\ref{overall_story}c. Note that interpretable reductions do not change the behavior of the agent, nor the control flow, but are only for improved visualization. The states remaining in the reduced diagram are \emph{decision points} (e.g. state $S_2$ is the single decision point in Figure~\ref{MM-samples}b), where the next state, and hence future behavior, depends on the observation. The decision points are the key states in the machine that dictate how its behavior is influenced by observations. It is these points where we can gain the most insight about an MM.

\begin{table*}[th]
\caption{MM results for control tasks and Atari environments. ``DP" refers to decision points, ``Obs."" is short for observations, and ``Perf." is short for performance.}
\begin{center}
\begin{tabular}{ c|c|c|c|c|c|c|c|c|c|c|c|c|c|c } 
\hline\hline
\multirow{2}{*}{Game} & \multicolumn{2}{c|}{QBN Sizes} & \multicolumn{4}{c|}{Original MM} & \multicolumn{4}{c|}{Functional pruning} & \multicolumn{4}{c}{Minimal MM} \\
\cline{2-15}
&\thead{$N_h$} & \thead{$N_o$}&\thead{DP} &\thead{States} & \thead{Obs.} & \thead{Perf.} & \thead{DP} & \thead{States} & \thead{Obs.} & \thead{Perf.}&\thead{DP} &\thead{States} & \thead{Obs.} & \thead{Perf.} \\
\hline\hline
\multirow{2}{*}{Acrobot}
& 4 & 4 & 9 & 12 & 38 & -77.1 & 2 & 4 & 3 & -80.7 & 3 & 3 & 11 & -80 \\
& 8 & 8 & 79 & 42 & 200 & -77.1 & 2 & 4 & 5 & -79.9 & 3 & 3 & 7 & -86\\
\hline
\multirow{2}{*}{CartPole}
& 4 & 4 & 6 & 7 & 18 & 500 & 2 & 4 & 5 & \textbf{500} & 2 & 3 & 8 & 500 \\
& 8 & 8 & 7 & 8 & 58 & 500 & 2 & 4 & 5 & \textbf{500} & 3 & 3 & 11 & 500 \\
\hline
\multirow{2}{*}{\shortstack{Lunar\\Lander}}
& 32 & 32 & 195 & 1426 & 1150 & 180.4 & 184 & 1387 & 980 & 172.2 & 41 & 41 & 92 & 165\\
& 32 & 64 & 249 & 1389 & 1437 & 204.9 & 230 & 1321 & 1327 & 197.3 & 19 & 19 & 73 & 147\\
\hline\hline
\multirow{2}{*}{Bowling}
& 32 & 50 & 5 & 608 & 525 & 60 & \textbf{0} & 530 & 437 & \textbf{60} & 5 & 5 & 19 & 60\\
& 64 & 100 & 5 & 630 & 552 & 60 & \textbf{0} & 546 & 488 & \textbf{60} & 3 & 4 & 12 & 60\\
\hline
\multirow{2}{*}{Boxing}
& 32 & 50 & \textbf{0} & 1274 & 1270 & 100 & \textbf{0} & 1274 & 1270 & \textbf{100} & 19 & 19 & 109 & 100 \\
& 64 & 100 & \textbf{0} & 1097 & 1095 & 100 & \textbf{0} & 1097 & 1095 & \textbf{100} & 14 & 14 & 101 & 100 \\
\hline
\multirow{2}{*}{Breakout}
& 32 & 50 & 4 & 2479 & 2466 & 404 & \textbf{0} & 2365 & 2345 & \textbf{404} & 8 & 8 & 28 & 404\\
& 64 & 100 & 5 & 1659 & 1608 & 404 & \textbf{0} & 1598& 1540 & \textbf{404} & 11 & 11 & 43 & 404\\
\hline
\multirow{2}{*}{Pacman}
& 32 & 50 & 43 & 728 & 716 & 3060 & \textbf{0} & 612 & 597 & \textbf{3060} & 21 & 21 & 70 & 3060\\
& 64 & 100 & 34 & 895 & 876 & 3060 & \textbf{0} & 776 & 758 & \textbf{3080} & 9 & 9 & 45 & 3060\\
\hline
\multirow{2}{*}{Pong}
& 32 & 50 & \textbf{0} & 383 & 369 & 21 & \textbf{0} & 383 & 369 & \textbf{21} & 3 & 3 & 12 & 21\\
& 64 & 100 & 2 & 384 & 369 & 21 & \textbf{0} & 271 & 268 & \textbf{21} & 4 & 4 & 10 & 21\\
\hline
\multirow{2}{*}{SeaQuest}
& 32 & 50 & 46 & 2167 & 2233 & 2580 & \textbf{0} & 1679 & 1577 & \textbf{2580} & 16 & 16 & 140 & 2580\\
& 64 & 100 & 18 & 2244 & 2261 & 2580 & \textbf{0} & 1834 & 1883 & \textbf{2580} & 17 & 17 & 135 & 2580\\
\hline
\multirow{2}{*}{\shortstack{Space\\Invaders}}
& 32 & 50 & 102 & 1700 & 1709 & 1820 & \textbf{0} & 1314 & 1350 & \textbf{1820} & 30 & 30 & 40 & 1820\\
& 64 & 100 & 35 & 1914 & 1852 & 1820 & \textbf{0} & 1832 & 1802 & \textbf{1820} & 11 & 11 & 27 & 1820\\
\hline\hline
\end{tabular}
\label{table:atari-results}
\end{center}
\end{table*}

\subsection{Differential Attention for Decision Points}\label{sec:attention}
Given an MM decision point, we are interested in understanding how the raw observations (e.g. image pixels) influence its decisions. In particular, for a pair of outgoing branches labeled by discrete observation $\hat{f}_1$ and $\hat{f}_2$, we would like to answer the question: \emph{``What features of the raw observations are most influential to selecting the $\hat{f}_1$ branch versus $\hat{f}_2$?"}. To help answer this, we consider pairs of raw observations $o_1$ and $o_2$ that occur at the decision point during MM execution (e.g. $o_{342}$ and $o_{267}$ in Figure~\ref{overall_story}c), such that $\hat{f}_i = E_o(o_i)$ where $E_o$ is the encoder of the observation QBN $Q_o$ which discretizes the input features. That is, $o_1$ and $o_2$ are example observations that cause the machine to differentiate between the branches. We then produce a \emph{differential attention map} $S(o_1, o_2)$ that highlights the parts of $o_1$ and $o_2$ that are most responsible for preferring branch $\hat{f}_1$ over $\hat{f}_2$. To compute $S(o_1, o_2)$, we focus on the set $F(o_1, o_2)$ of discrete features produced by $E_o$ that differ between $o_1$ and $o_2$. As described below, for each $f \in F(o_1,o_2)$, we first produce an attention map $S[f](o_1, o_2)$ that highlights the parts of $o_1$ and $o_2$ that ``explain" the difference in value of $f$. $S(o_1, o_2)$ is then just the average of the individual maps.

We compute each $S[f](o_1, o_2)$ via a straightforward, but novel, adaptation of the \emph{Integrated Gradient (IG)} attention approach \cite{CVXAIsundararajan2017axiomatic}. Originally, IG was used to compute attention maps that explain the decision of a classifier on a single image/observation $o$. Fundamental to the approach is the notion of a \emph{baseline} image $o_b$, which is used as a reference that is assumed to not excite the classifier. In an image domain, the baseline is often a constant or noise image. Let $f(o)$ be the classifier response for observation $o$ (usually the largest class-specific input to the softmax layer), noting that $f(o_b)$ will be small, and let $IG_i[f](o,o_b)$ be the corresponding attention value produced by IG for feature/pixel $i$. The key property of IG, which makes it a meaningful notion of attention, is the relation $\sum_i IG_i[f](o,o_b) = f(o)-f(o_b)$. Thus, the attention value of pixel $i$ can be viewed as its additive contribution to the difference in classification responses for $O$ over $O_b$. For space reasons we refer the reader to the original paper \cite{CVXAIsundararajan2017axiomatic} for details of the IG computation.

\begin{figure*}[t]
\begin{center}
\subfigure[]{
\includegraphics[width=0.27\textwidth]{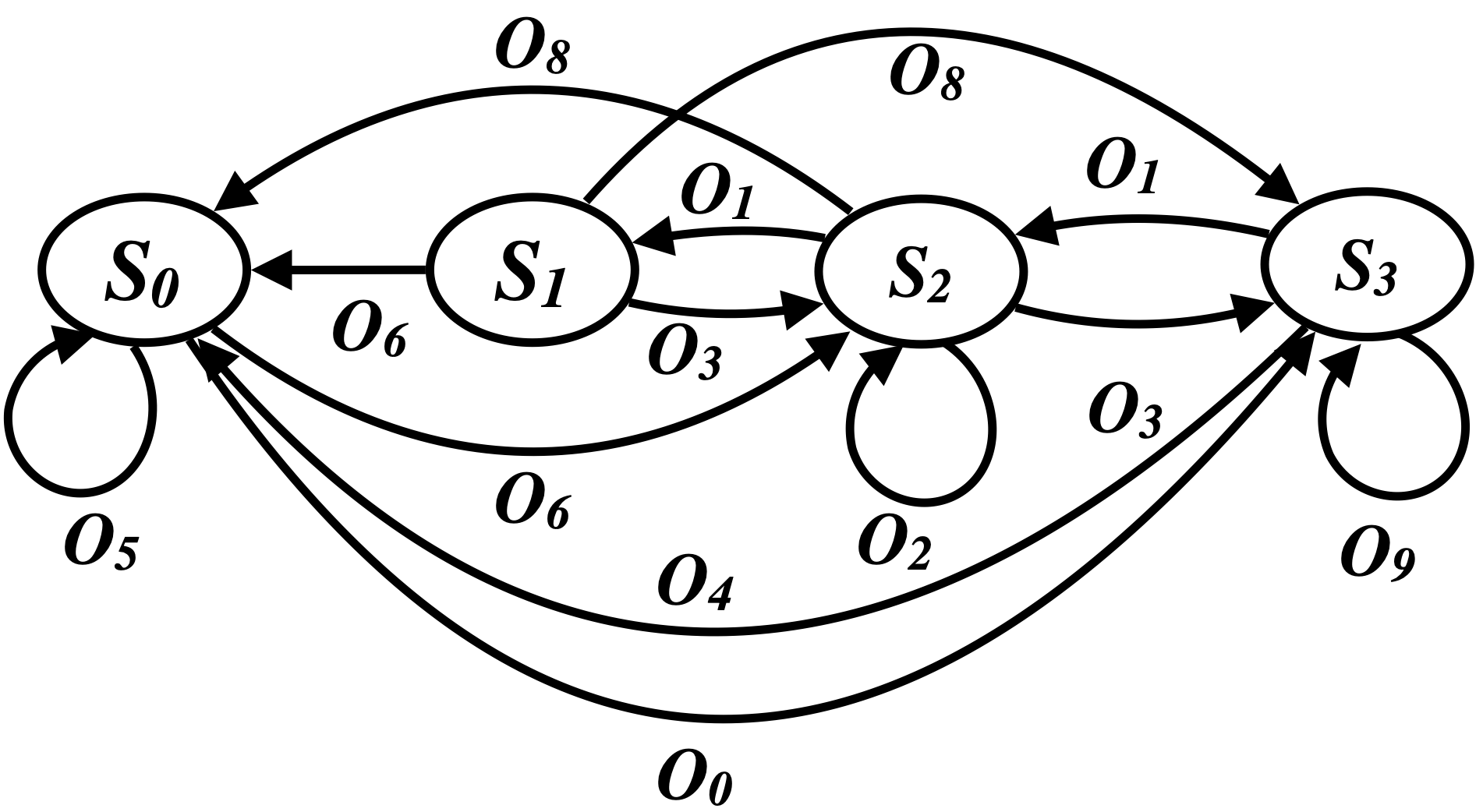}}
\subfigure[]{
\includegraphics[width=0.27\textwidth]{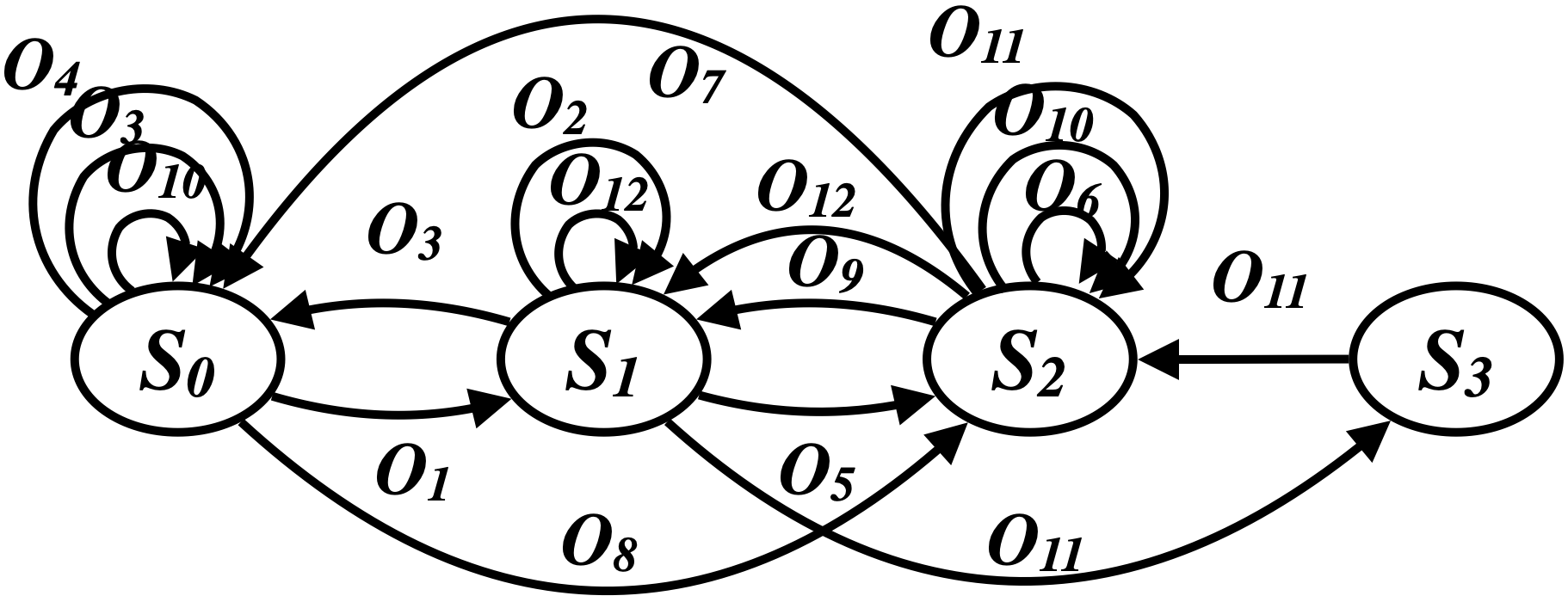}}
\subfigure[]{
\includegraphics[width=0.06\textwidth]{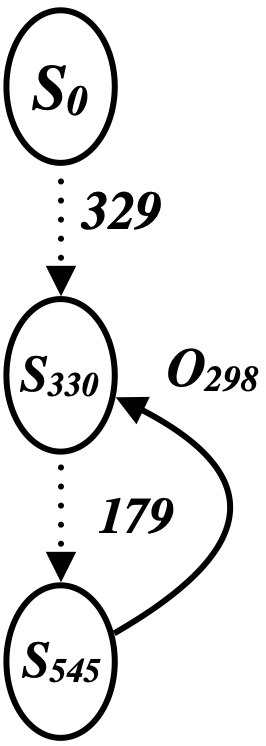}}
\subfigure[]{
\includegraphics[width=0.27\textwidth]{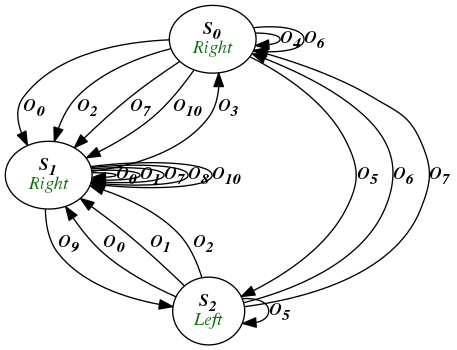}}
\end{center}
\vspace{-1em}
\caption{a) Pong minimal MM, b) Bowling minimal MM, c) Bowling pruned MM, d) Acrobot minimal MM.}
\label{minimalMM}
\vspace{-1em}
\end{figure*}

Figure~\ref{fig:attention} shows how we adapt IG to compute a differential attention map $S[f](o_1, o_2)$ by treating $o_2$ as the baseline and letting the response function $f \in F(o_1,o_2)$ be the continuous features computed by the encoder just before the discretization step. That is, differential attention is given by $S[f](o_1,o_2)= IG[f](o_1,o_2)$. This means that differential attention satisfies $\sum_i S_i[f](o_1,o_2) = f(o_1) - f(o_2)$, which has the interpretation that each attention value $S_j[f]$ can be viewed as an additive contribution to the difference in response for $o_1$ and $o_2$, i.e. the preference of $o_1$ over $o_2$. \footnote{Source code is available at: \href{https://github.com/modanesh/Differential\_IG}{github.com/modanesh/Differential\_IG}}

\subsection{Functional Pruning}\label{sec:functional}
After exploring the reduced MM graphs and attentions in Atari, we made two high-level observations. First, as shown in our experiments the differential attention results often indicated that observations were not being used in a strategically meaningful way, e.g., branching on observations that were extremely similar. Second, the graphs beyond different branches at a decision point were often very similar and appeared to address similar situations. We hypothesized that many decision points may not be strategically meaningful, but rather just an artifact of learning. That is, even if a branching decision is not required at a certain point in a game, the inclusion of a decision point that conditions on an arbitrary part of the observation will not hurt the performance as long as good behavior is learned for each of the resulting branches. In such situations, the choice of which branch to traverse may be arbitrary and any one of them may work. Note that the usual attention-based tools \cite{pmlr-v80-greydanus18a, DBLP:journals/corr/abs-1809-06061, gupta2020explain} will simply indicate the agent's attention and can lead a human to incorrectly infer strategic relevance. 

Detecting unnecessary decision points cannot be done by just graph analysis---rather, empirical analysis of modified machines is required. To do this, we conduct a simple form of \emph{functional pruning} (details and pseudo-code in appendix), to identify and prune unnecessary branches at decision points. Our approach considers each MM decision point, and prunes each of the branches one at a time, in order of the least to most frequently visited based on multiple MM runs. When a branch is pruned, we empirically estimate the performance of the resulting MM, noting that when the machine would have previously taken the pruned branch, we force it to take the most frequent branch. If the empirical performance does not degrade beyond a threshold, we permanently remove the branch and move on to the next pruning step. \emph{The intent of function pruning is not to preserve logical equivalence.} Rather, the intent is to test whether observations were strategically important or just an artifact of learning. After gaining the insights, one could decide to use the original machine or attempt to improve it.

We found greedy pruning to be effective for the MMs considered in this paper. For example, in Figure~\ref{overall_story} we see that functional pruning resulted in pruning all but one branch from the single decision point in the MM, leaving an open-loop policy. In such cases, when an MM can be functionally pruned to the point of removing all decision points and leaving an open loop policy, we say that the MM is a \emph{pruned open-loop policy}, which generalizes the notion of open-loop policy to include machines that condition on observations, but in ways that are not strategically necessary. As an analogy, consider a pruned open-loop policy for a human driving to a store along one of two equally good routes. The human may measure the temperature and decide between the routes depending on whether the temperature was an even or odd value. The observation had no real impact on the quality of the policy, since a fixed route could have been selected, but behavior is still seen to depend on observations.  

\section{Experiments}
The only prior approach to compare against is the recent MM minimization approach for analysis of MMs \cite{MISCkoul2018learning}. As described earlier, we have found it very difficult to gain insights from minimized MMs. To further illustrate this point, in the appendix we provide a quantitative and qualitative comparison of the MM approach to our approach in Atari games. Overall, the minimization approach does not reveal the insights of our new approach. 

We consider 7 deterministic Atari environments: Bowling, Boxing, Breakout, MsPacman, Pong, SeaQuest and SpaceInvaders, and 3 stochastic discrete-action classic control tasks: Acrobot, CartPole, and LunarLander. For each experiment, we follow the choices of prior discretization work \cite{MISCkoul2018learning} for pre-processing, RPN architecture, QBN architectures, and training via A3C \cite{MISCmnih2016asynchronous} reinforcement learning. Detailed information on these choices along with hyperparameter choices are in the appendix. In addition to A3C, we investigate and interpret another policy learned by the R2D2 algorithm \cite{kapturowski2018recurrent} which results are provided in the appendix. We considered two sets of QBN sizes for each environment, shown in Table~\ref{table:atari-results}. From the table we see that the agent performance remains the same after reductions for all domains, except for Acrobot and LunarLander, where functional pruning reduced performance within the specified tolerance. The table also gives the number of states and observations, $N_h$ and $N_o$, after the reductions. Interestingly, in Atari games, no decision points are left after functional pruning, i.e. all of the policies were ``pruned open-loop" policies. The following case studies illustrate how our approach is useful for gaining insights and revealing unexpected properties of the decision logic. Additional examples are in the Appendix. 

\subsection{Atari: Case Studies}\label{subsec:case_studies}
\textbf{Example Insights: Pong and Bowling.} The RPN for Pong achieved a maximum score of 21. The policy displays repetitive behavior by performing a ``kill shot" against the opponent to win each point, though the behavior is not exactly the same across all shots. Figure~\ref{overall_story}b shows a view of the original large MM. Figure~\ref{overall_story}c shows the graph after the interpretable reductions, which is quite small with only one decision point at state $S_{197}$. 
This key decision point is the starting point of 3 possible loops, depending on the branch taken, and is entered once per round (each round is one point). This raised the questions of \emph{``What basis is the machine using to decide which loop to enter?"} and \emph{``Is there a strategic reason for the branching decision?"}.  

To investigate, we computed the differential attention for the decision of choosing $S_{275}$ over $S_{352}$, as shown in Figure~\ref{fig:attention}. It is striking that the sample observations associated with the branches are almost identical. The differential attention indicates that the ball region and tips of the paddles are the most critical factors in deciding between the branches. Close inspection reveals that the appearance and location of the ball in the two observations are subtly different. To understand this, we observed that these differences are due to the fact that at the beginning of each round the starting position of the ball is minutely different for even versus odd rounds, which translated to the small difference observed at the decision point. Intuitively, this difference did not appear to have a strategic value. Indeed, after functional pruning (Figure~\ref{overall_story}d), we see that all branches were removed except for the one through $S_{275}$, leaving an open-loop strategy with no loss in performance. The even versus odd branching, was an unnecessary artifact of the RNN learning process. 

For Bowling, the original large MM had 630 discrete states and 552 discrete observations. Our interpretable reductions revealed only 5 of those states corresponded to true decision points. Further, Figure~\ref{minimalMM}c shows the result of functional pruning to get an MM with no loss in performance. Similar to Pong we end up with a pruned open-loop policy. Again the observations used at decision points were not strategically relevant and rather artifacts of learning.

\begin{figure}[t]
\begin{center}
\subfigure[]{
\includegraphics[width=0.11\columnwidth]{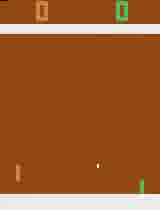}
\includegraphics[width=0.11\columnwidth]{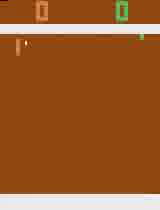}
\includegraphics[width=0.11\columnwidth]{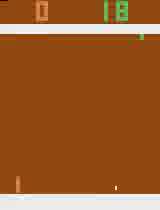}
\includegraphics[width=0.11\columnwidth]{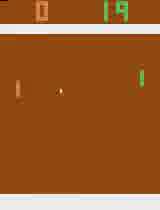}}
\subfigure[]{
\includegraphics[width=0.11\columnwidth]{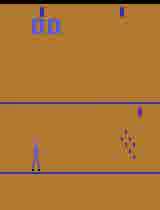}
\includegraphics[width=0.11\columnwidth]{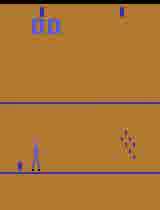}
\includegraphics[width=0.11\columnwidth]{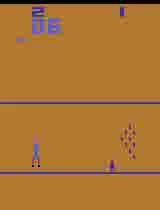}
\includegraphics[width=0.11\columnwidth]{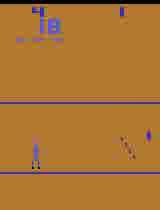}}
\end{center}
\vspace{-1em}
\caption{Four observations that enter a state in the corresponding minimal MM. a) Frames of Pong for in-going observations to $S_2$, b) Frames of Bowling for in-going observations to $S_0$.}
\label{frames}
\vspace{-1em}
\end{figure}

{\bf Comparison to Prior Work.} We now compare to the minimization approach of \cite{MISCkoul2018learning}. For Pong, our approach was able to isolate a single decision point (Figure~\ref{overall_story}d) where behavior depended on observations in an understandable way, which was ultimately determined to be non-strategic. Meanwhile, Figure~\ref{minimalMM}a shows the minimal MM produced by prior work for the same initial MM. This minimal MM merges the key decision point with semantically unrelated states (e.g. from multiple macros), obscures insights, and gives no understanding of how memory and observations are used. To highlight this, Figure~\ref{frames}a demonstrates four frames of Pong which enter $S_2$ in the corresponding minimal MM. This is an evidence of minimal MMs merging states that are semantically distinct. Further, it is unclear how the equivalent of functional pruning could be done using the minimal MM, due to the merging of decision points. Also, it is unclear how to uncover the key insight identified by our approach by starting with the minimal MM. 

A similar comparison holds for Bowling where our approach resulted in the open-loop policy (Figure~\ref{minimalMM}c). Rather, the minimal MM of the same initial policy is shown in Figure \ref{minimalMM}b, which resembles the tightly coupled minimal MM of Pong. Figure \ref{frames}b shows frames that lead to decision point $S_0$, which appear to be semantically very different from a human perspective. This is also the case for other states, which makes it very difficult to extract meaningful insights from the minimal MM. Again, it is completely unclear how we could start with the minimal machine and gain the realization that the observations play no significant strategic role, which our approach revealed. 

\textbf{Overall Results.} Due to page limit, qualitative comparisons of other Atari games are provided in the appendix. Nevertheless, they provide similar distinctions which show our approach's advantage over \cite{MISCkoul2018learning}'s. Table~\ref{table:atari-results}, gives information about the MMs before and after functional pruning for each game, and a comparison with the minimization approach. Note that 0 decision points indicates an open-loop policy. We see that before functional pruning there is only one case of open-loop policies: Boxing. All other MMs have at least one decision point. This initially makes one to believe that observations are a key part of the overall MM strategy. However, in each such case, we found that it was rare to find a decision point where observations provide strategic values at a decision point. This was confirmed by our most striking finding. After functional pruning, each of the games resulted in open-loop MMs (i.e. zero decision points). Thus, in all cases, Atari RPNs produced MMs that were pruned open-loop policies.  

\begin{figure}[t]
\begin{center}
\subfigure[]{
\includegraphics[width=\columnwidth]{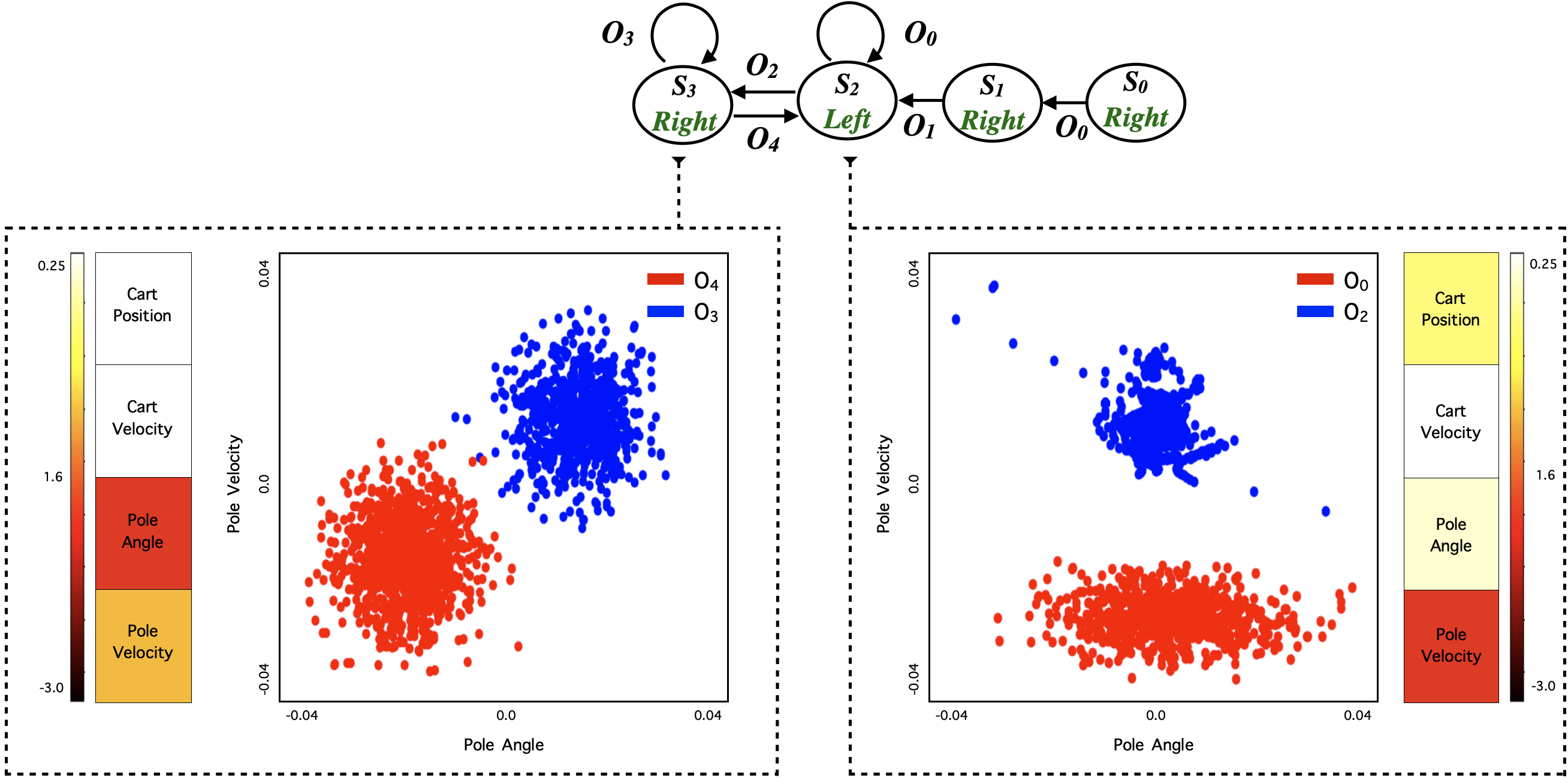}}
\subfigure[]{
\includegraphics[width=\columnwidth]{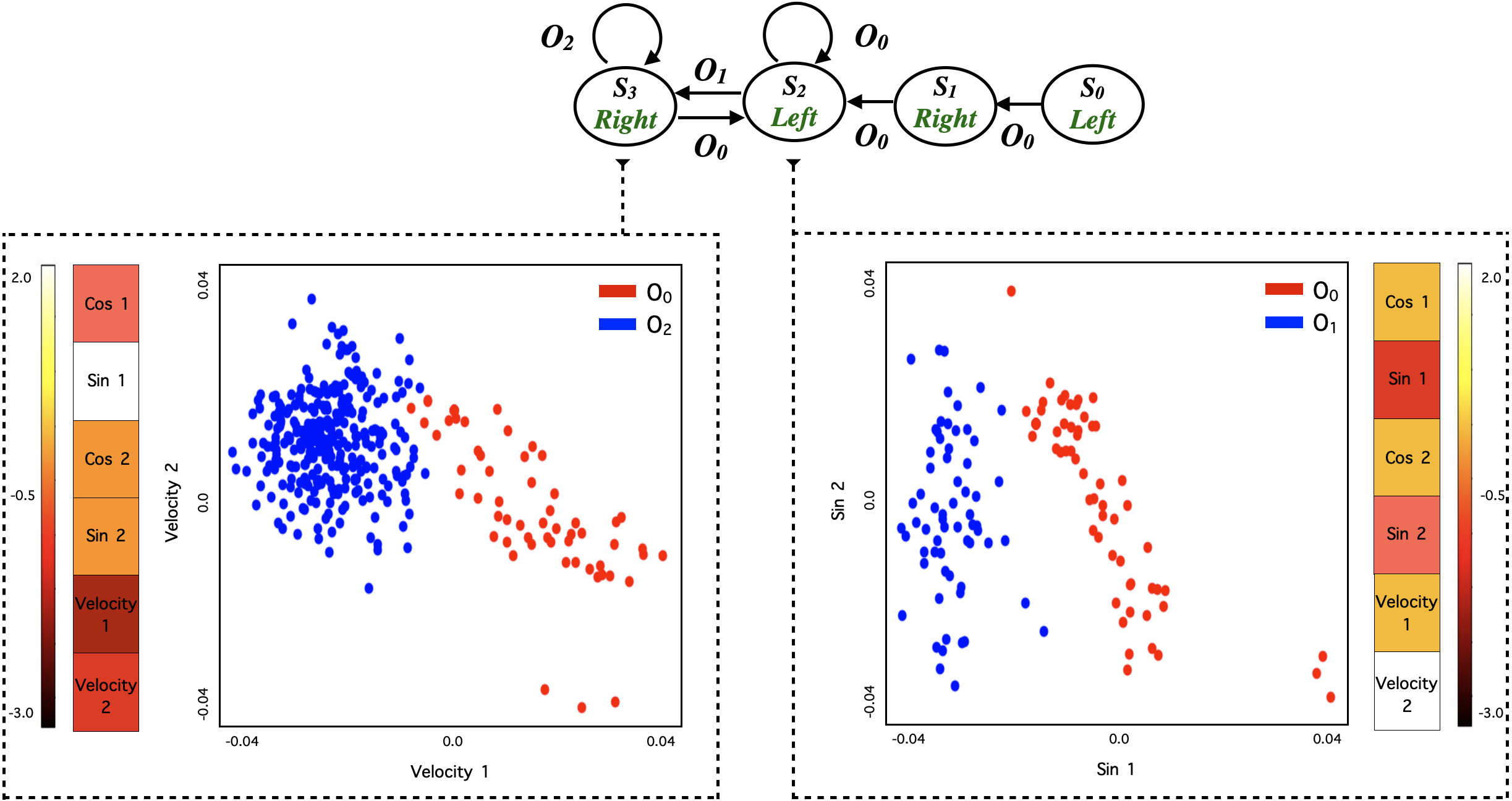}}
\end{center}
\vspace{-1em}
\caption{Pruned MMs for control tasks. a) CartPole, and b) Acrobot. In each case, we show attention of features for decision points. Also, we show scatter plots of continuous observations during an episode at each decision point for the two most salient features, where color indicates the discrete machine observation.}
\label{fig:control_pruned_machine}
\vspace{-1.5em}
\end{figure}

\subsection{Stochastic Classic Control Tasks}
\textbf{Acrobot.} This control task includes two joints and two links, where the joint between the two links is actuated. Initially, the links are hanging downwards, and the goal is to swing the end of the lower link up to a given height. The state vector gives the $\sin$ and $\cos$ of the two rotational joint angles and the joint angular velocities. The actions involved are -1,0 or +1 torque. We share pruned state machine for QBN sizes (4, 4) in Figure~\ref{fig:control_pruned_machine}b. We applied our differential attention approach to decision point $S_2$ and found that the most important observation features were `sin of joint angle 1" and ``joint 2 velocity". The Figure~\ref{fig:control_pruned_machine}b shows the scatter plot of the decisions at $S_2$ versus these features. The plot reveals that for positive $\sin$ values, a torque of -1 is applied by starting in $S_2$ forcing link 1 and link 2 to rise against gravity. This torque is applied until it transitions to $S_3$ via $O_1$ which corresponds to a positive velocity for joint 2 as shown in the scatter plot. This happens when link 2 cannot go further up against gravity under the current momentum. In this state, it applies +1 torque to supplement the momentum provided by the pull of gravity. It transitions back to $S_2$ via $O_0$ only when the joint velocity of link 1 is positive, indicating that it cannot go further up. This loop between state $S_2$ and $S_3$ generates enough momentum to eventually reach the goal. 

In contrast, we compare against the minimal MM (Figure~\ref{minimalMM}d) extracted via prior work \cite{MISCkoul2018learning}. This MM, is fully connected and merges many different types of observations. Through various previous attempts, we were unable to elicit a clear description of the machines operation. Rather, our reduction approach was able to lead to the above relatively clear understanding of the machine. 

\textbf{CartPole.} We use the standard OpenAI CartPole environment with randomized initial states. 
For QBN pairs of (4,4), Table~\ref{table:atari-results} shows that the number of decision points is 6 before functional pruning and reduces to 2 after. The pruned MM is shown in Figure~\ref{fig:control_pruned_machine}a, which has the same simple structure as for Acrobot. The machine primarily transitions between $S_2$ (left movement) and $S_3$ (right movement) with self-loops at those states between transitions. The figure shows the differential attention computed over the features at $S_2$ and $S_3$ with the key features being ``pole-velocity" followed by ``pole-angle". The scatter plots for the decisions against these features show that at $S_2$ there is a clear threshold of positive pole-velocity that transitions to $S_3$ to take the ``right" action, and otherwise continues with ``left". This is an intuitive strategy for reducing the velocity. Similar insights are gained via the scatter plot at $S_3$, but here both pole-velocity and pole-angle play a role in the decision. Again, our approach was able to produce an MM that has meaningful interpretation. 


\textbf{Lunar Lander.} This task involves landing a rover using actions that fire one of three thrusters: main, left, right, and no-op. The MMs for both QBN sizes were significantly larger than for the above tasks, likely due to increased complexity and stochasticity. Functional pruning was ineffective, leaving 184 and 230 decision points for the two QBN sizes, which indicates the observations have a strong strategic role. We analyzed many of the decision points and found that the ``distance to the ground" is usually the most salient feature across decision points. One exception is the initial decision point. At this decision point there were three branches with different observations leading to states with different fire actions \textit{main engine}, \textit{right engine} and \textit{left engine}. The attention comparison of these observations shows prime differences in x-velocity and y-velocity, thereby opposite direction engines are fired to stabilized the rover. This decision point does not attend to rover coordinates (x,y position) and leg-positions, which are more relevant when the lander is closer to the ground. While it is difficult to articulate a concise description of such a machine, we see that this analysis approach is able to provide insights about the decisions that may help build confidence or identify concerns.

\section{Discussion and Current Challenges}
Our analysis is the first to provide such detailed insights into the decision making of RPNs for deterministic Atari games. Indeed, the observation that all of the considered policies resulted in pruned open-loop controllers was unexpected apriori and not apparent from prior work. For example, prior work on attention analysis of Atari policies, even for the deterministic setting, leaves one to believe that observations play key roles in decision making. It is tempting to discount the above insights, due to the deterministic setting, since 1) it is clear that there exist open-loop controllers for any deterministic domain, and 2) prior work has shown that search is able to find effective open-loop plans for some Atari games \cite{bellemare2013arcade, machado2018revisiting}. However, these points do not imply that an RPN would necessarily learn an open-loop controller and indeed we observed that they do not. It is reasonable to expect that RPNs would meaningfully use observations to get a more general policy, but we instead saw the role for observations was very different. This demonstrates the importance of developing a variety of tools to reveal insights that rely less on human assumptions and interpretation.   

Our preliminary experience with larger and more complex environments shows that sometimes our approach does not reveal easily analyzable MMs. This was apparent in stochastic Atari experiments, where we observed that the discrete sequences produced across different episodes have some overlap, but are dominated by large numbers of disjoint states. Some potential explanations for this are: 1) An inadequacy of our approach--e.g., the quantization process and/or reduction steps/analysis  may need to be improved; 2) Our approach may be identifying the fundamental nature of the learned RNN policies. That is, the policies may effectively use large numbers of trajectories to encode large numbers of effectively (pruned) open-loop patterns. As new observations are encountered the machines then map to encoded patterns in a nearest neighbor style. This second possibility would suggest the need for improved RPN training approaches to support data efficiency and interpretability. 

\section*{Acknowledgements}
This material is based upon work supported by the Defense Advanced Research Projects Agency
(DARPA) under the XAI and CAML programs.  Any opinions, findings and conclusions or recommendations expressed in this material are those of the authors and do not necessarily reflect the views of the DARPA, the Army Research Office, or the United States government.

\bibliography{example_paper}
\bibliographystyle{icml2021}

\end{document}